%% file: main.tex
\pdfoutput=1

\documentclass[11pt]{article}

\usepackage{ACL2023}

\usepackage{times}
\usepackage{latexsym}

\usepackage[T1]{fontenc}

\usepackage[utf8]{inputenc}

\usepackage{microtype}
\definecolor{blue-green}{rgb}{0.0, 0.87, 0.87}

\definecolor{blue(pigment)}{rgb}{0.2, 0.2, 0.6}

\definecolor{brightcerulean}{rgb}{0.11, 0.67, 0.84}

\usepackage{inconsolata}

\usepackage{graphicx}
\usepackage{multirow}
\usepackage{tabularx}
\usepackage{enumitem}
\usepackage{booktabs, wrapfig}
\usepackage{subcaption}

\usepackage{pifont}

\title{A Deep Dive into the Trade-Offs of Parameter-Efficient \\ Preference Alignment Techniques}

\author{
Megh Thakkar$^{1, 2}$ \quad Quentin Fournier$^{1}$ \quad Matthew Riemer$^{1,2,3}$ \quad \textbf{Pin-Yu Chen}$^{3}$ \\ 
\textbf{Amal Zouaq}$^{1,4}$ \quad \textbf{Payel Das}$^3$ \quad \textbf{Sarath Chandar}$^{1,4,5}$ \\ \\
$^1$Mila – Quebec AI Institute \quad $^2$Université de Montréal \quad $^3$IBM Research \\
\quad $^4$Polytechnique Montréal \quad $^5$Canada CIFAR AI Chair \\
\texttt{
\{firstname.lastname\}@mila.quebec} \\  \quad \texttt{pin-yu.chen@ibm.com} \quad \texttt{daspa@us.ibm.com}}

\begin{document}

\maketitle

\begin{abstract}
	Large language models are first pre-trained on trillions of tokens and then instruction-tuned or aligned to specific preferences. While pre-training remains out of reach for most researchers due to the compute required, fine-tuning has become affordable thanks to parameter-efficient methods such as LoRA and QLoRA. Alignment is known to be sensitive to the many factors involved, including the quantity and quality of data, the alignment method, and the adapter rank. However, there has not yet been an extensive study of their effect on downstream performance. To address this gap, we conduct an in-depth investigation of the impact of popular choices for three crucial axes: (i) the alignment dataset (HH-RLHF and BeaverTails), (ii) the alignment technique (SFT and DPO), and (iii) the model (LLaMA-1, Vicuna-v1.3, Mistral-7b, and Mistral-7b-Instruct). Our extensive setup spanning over 300 experiments reveals consistent trends and unexpected findings. We observe how more informative data helps with preference alignment, cases where supervised fine-tuning outperforms preference optimization, and how aligning to a distinct preference boosts performance on downstream tasks. Through our in-depth analyses, we put forward key guidelines to help researchers perform more effective parameter-efficient LLM alignment.
\end{abstract}

\section{Introduction}

Large Language Models (LLMs) have achieved human-like performance across various tasks such as summarization, commonsense reasoning, and open-ended generation~\citep{zhao2023survey}. These LLMs have billions of parameters and are \textit{pre-trained} on trillions of tokens scraped from the web. A lucrative utilization of LLMs is in the form of autonomous agents, to make them follow user instructions and adhere to specific preference requirements~\citep{wang2023survey}. However, the pre-trained models are often incapable of following instructions, and they need to be \textit{aligned} using specially curated preference alignment datasets and methods for generalization~\citep{Mishra2021CrossTaskGV}.

Alignment methods either involve fine-tuning the pre-trained model using auto-regressive language modeling over the ground truth completions (supervised fine-tuning or SFT)~\citep{alpaca} or using specialized alignment methods such as reinforcement learning from human feedback (RLHF)~\citep{christiano2023deep}, direct preference optimization (DPO)~\citep{rafailov2023direct}, or prompt tuning~\citep{xue2023parameterefficient}. However, applying these methods to the full models is computationally expensive due to their large sizes. Parameter-efficient training (PEFT) methods such as Low-Rank Adaptation (LoRA)~\citep{hu2022lora} and QLoRA~\citep{dettmers2023qlora} have achieved comparable performance to full fine-tuning of LLMs at a much lower cost. This has enabled researchers to experiment with preference alignment datasets, methods, and models on systems with a single GPU. However, alignment is sensitive to numerous factors and design choices involved in the training~\citep{wang2023aligning}.

\begin{table*}[!htb]
	\centering
	\small
	\begin{tabular}{p{.225\linewidth}p{.225\linewidth}p{.45\linewidth}}
		\toprule
		\textbf{Existing works}                                                                &   
		\textbf{Limitations}                                                                   &   
		\textbf{This work}
		\\\midrule
		DPO is better at alignment than SFT~\citep{rafailov2023direct}                         &   
		Evaluated with full model training of instruction-tuned models on limited NLP tasks    &   
		We compare SFT and DPO over distinct preferences using pre-trained and instruction-tuned models and find that 
		\textit{DPO is suited for instruction-tuned models, but SFT is suited for pre-trained models}
		\\ \midrule
		Mixing two preferences for preference alignment has trade-offs~\citep{bai2022training} &   
		Evaluated with RLHF in full fine-tuning settings on limited tasks                      &   
		We analyze trade-offs for SFT and DPO across different models and preferences and observe that
		\textit{mix of preferences leads to degradation for both SFT and DPO when using PEFT}
		\\ \midrule
		Aligning to a preference improves its performance~\citep{bai2022training}              &   
		Evaluated with full fine-tuning on limited NLP tasks                                   &   
		We experiment across models and alignment methods with PEFT 
		and see that \textit{often aligning to distinct preferences leads to improvements and aligning to same preference leads to degradation} 
		\\ \midrule
		No existing works                                                                    &   
		&   
		We analyze the effects of the number of samples used for alignment and observe that
		\textit{SFT decreases the performance for instruction-tuned models, while DPO improves or obtains performs similar for instruction-tuned and pre-trained models}  
		\\ \midrule
		No existing works &   &   
		We analyze the effects of merging models trained with different alignment methods and preferences and observe that
		\textit{merging models leads to improvement over individual models aligned to distinct preferences}
		\\ \bottomrule
	\end{tabular}
	\caption{Comparing some of our experiments that address the limitations of existing works.}
	\label{tab:findings}
\end{table*}

The design choices to align LLMs fit into one of the following three broad, crucial axes: (i) the quality and quantity of the alignment \textit{dataset}, (ii) the preference alignment \textit{method}, and (iii) the nature of the base \textit{model}. Given the increasing interest in preference alignment, an in-depth analysis of the effect of these axes on downstream performance is required. To the best of our knowledge, no extensive study has investigated them, especially in a PEFT setting. We fill this important gap by attempting to answer various key research questions across these three axes:

\noindent \textbf{Alignment Dataset} How do the informativeness and quality, number of samples, and content of the preference dataset impact the downstream performance?

\noindent \textbf{Alignment Method} How do different alignment methods affect pre-trained and instruction-tuned models over complementary preferences?

\noindent \textbf{Nature of the Base Model} How does the downstream performance compare between pre-trained models, instruction-tuned models, and their merged variants?

Though our study covers all three axes, given the rapidly growing number of options for each, we restrict the studied choices to the most popular ones. Specifically, we perform our experiments on two commonly used preferences in literature to study alignment trade-offs: harmlessness and helpfulness. We use the (i) two most popular preference alignment datasets with harmlessness and helpfulness annotations: HH-RLHF~\citep{bai2022training} and BeaverTails~\citep{ji2023beavertails}, with (ii) the two most widely-used alignment methods: SFT~\citep{alpaca} and DPO~\citep{rafailov2023direct}, and (iii) two commonly used LLMs, LLaMA-1~\citep{touvron2023llama} and Mistral-7b~\citep{jiang2023mistral} along with their instruction-tuned versions, Vicuna-v1.3~\citep{vicuna2023} and Mistral-7b-Instruct. For an in-depth study of PEFT methods, we conducted all experiments using both LoRA and QLoRA. Our extensive analysis across these core axes reveals certain consistent trends and unexpected findings, as shown in Table~\ref{tab:findings}. We hope that consolidating the key findings into guidelines will benefit the community in conducting impactful research toward LLM alignment. Our contributions can be summarized as:
\begin{itemize}[itemsep=2pt,parsep=2pt,partopsep=2pt,leftmargin=*,topsep=2pt]
	\item We provide an in-depth study into the trade-offs of parameter-efficient preference alignment training, particularly when using LoRA and QLoRA.
	\item We conduct over 300 experiments across three core axes of preference alignment: the dataset, the alignment method, and the model.
	\item Through experiments on $5$ evaluation benchmarks across harmlessness and helpfulness, we consolidate our key findings as guidelines for more effective preference alignment practices.
\end{itemize}

\section{Background}

\paragraph{Alignment Methods} reduce the mismatch between an LLM's pre-training and preference requirements of users, as LLMs are not pre-trained on human preference objectives. We describe the two most widely used alignment methods.

Supervised Fine-Tuning (SFT) uses a pair of input instructions and corresponding outputs to fine-tune the LLM using autoregressive language modeling. SFT is often used at the ``instruction-tuning'' stage of models such as Alpaca~\citep{alpaca} and Mistral-Instruct~\citep{jiang2023mistral}.

Direct Preference Optimization (DPO)~\citep{rafailov2023direct} is a stable and optimized alternative to reinforcement learning algorithms such as RLHF~\citep{christiano2023deep}. These methods have been used to align LLMs better using paired human preference data: data consisting of accepted and rejected outputs given an instruction.

\paragraph{Parameter-Efficient Training (PEFT)} methods require updating only a fraction of the parameters of the full model while achieving performance close to that of full fine-tuning. Examples of these methods include adapters~\citep{houlsby2019parameter}, prefix-tuning~\citep{li-liang-2021-prefix}, and prompt-tuning~\citep{lester-etal-2021-power}. Low-Rank Adaptation (LoRA)~\citep{hu2022lora} is another popular method that inserts a smaller number of new weights into the model, the only trainable parameters. QLoRA~\citep{dettmers2023qlora} improves the efficiency of LoRA by reducing the memory footprint and storage requirements for training while getting similar performance.

\section{Experimental Setup}

\paragraph{Preferences and Preference Alignment Datasets} We focus on two distinct preferences commonly used to study performance trade-offs in alignment: harmlessness and helpfulness. When a model is aligned on a preference, there is often a performance trade-off on complementary preferences~\citep{bai2022training}. Hence, these preferences enable us to study these trade-offs in-depth across multiple axes. We use the two most popular datasets for these preferences: HH-RLHF~\citep{bai2022training} and BeaverTails~\citep{ji2023beavertails}. HH-RLHF has explicit splits for harmlessness and helpfulness, and each dataset sample consists of a prompt, a chosen sample, and a rejected sample. However, many ``harmless'' responses in the dataset involve the model refraining from giving a response altogether~\citep{ji2023beavertails}. BeaverTails addresses this issue by providing more informative and elaborate responses. Each dataset sample consists of a prompt, two responses, a safety label for each response, and a label for the preferred response.

\paragraph{Base Models} We investigate the preference alignment trade-offs using pre-trained and instruction-tuned models. We use the two most popular 7 billion parameter pre-trained models, LLaMA-1~\citep{touvron2023llama} and Mistral-7b~\citep{jiang2023mistral}. For completeness, we use their instruction-tuned counterparts, Vicuna-v1.3~\citep{vicuna2023} and Mistral-7b-Instruct~\citep{jiang2023mistral}. The models trained on the harmless preferences are denoted by the suffix ``-Harmless'' (e.g., Mistral-7b-Harmless), and those trained on the helpful preferences are denoted by ``-Helpful''.

\begin{figure*}[!htb]
	\centering
	\includegraphics[width=\textwidth]{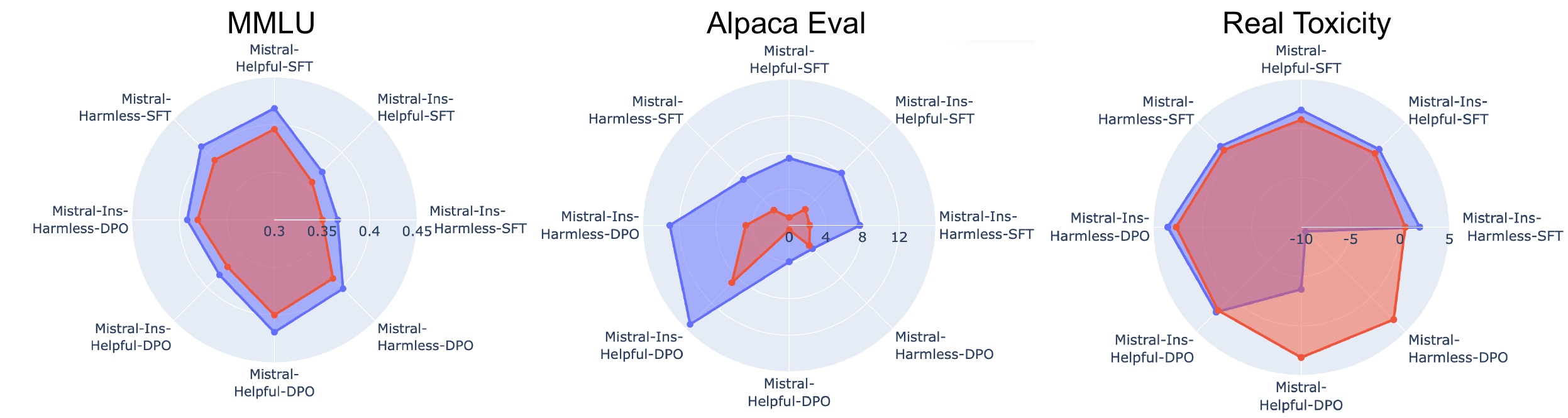}
	\caption{\label{fig:hh_vs_pb} Performance comparison for helpful and harmless benchmarks when models are aligned using QLoRA over HH-RLHF (in \textbf{\colorbox{red!30}{red}}) and BeaverTails (in \textbf{\colorbox{blue!30}{blue}}). We observe better performance when using a more informative and high-quality preference alignment dataset, albeit it is often overfitting for non-instruction tuned models when aligned using DPO (Section~\ref{subsec:hh_vs_bv}).}
\end{figure*}

\paragraph{Evaluation Tasks and Metrics} For evaluating helpfulness, we use (i) MMLU~\citep{hendrycks2021measuring}, a dataset consisting of 57 tasks to test the knowledge acquired by the models, (ii) Big-bench hard (BBH)~\citep{suzgun2022challenging}, a diverse evaluation suite of 23 tasks, and (iii) Alpaca Eval~\citep{alpaca_eval}, a human-curated suite of 805 questions across different tasks evaluated against GPT-4. For harmlessness evaluation, we use (i) RealToxicity~\citep{gehman-etal-2020-realtoxicityprompts}, a prompt dataset to measure toxic generation in language models. We choose the $1000$ most severe prompts from the dataset, and (ii) Red-Instruct's~\citep{bhardwaj2023redteaming} DangerousQA with chain-of-utterance, a dataset comprising 200 harmful questions across six adjectives—racist, stereotypical, sexist, illegal, toxic, and harmful. Following existing work, we evaluate MMLU using accuracy, BBH using exact match, and Alpaca-Eval using win rate against GPT-4. For RealToxicity, we use the score given by a reward model trained to classify toxic generations\footnote{\href{https://huggingface.co/nicholasKluge/ToxicityModel}{https://huggingface.co/nicholasKluge/ToxicityModel}}, and 100-Attack Success Rate for Red-Instruct.

\paragraph{Training Setup} We experiment with both QLoRA and LoRA for alignment training. For SFT, we keep the rank of the LoRA/QLoRA matrix as $64$ and LoRA/QLoRA alpha as $16$. For DPO, we keep the rank of the LoRA/QLoRA matrix as $16$ and LoRA/QLoRA alpha as $32$. We use a batch size of $16$, a learning rate of $2e-4$ for SFT and $5e-5$ for DPO, and train the models for $700$ steps. We perform all our experiments on a single $40$GB A100 GPU and run 5 seeds for all experiments.

\section{Effects of the Alignment Dataset}
\label{sec:alignment_dataset}

We investigate the effects of the alignment dataset in terms of its informativeness and quality, the number of samples used, and the preference sets and mixtures used for alignment.

\subsection{Impact of Quality}
\label{subsec:hh_vs_bv}

Previous works have shown that more informative and high-quality responses improve the alignment of models~\citep{ji2023beavertails}. However, this has only been explored for instruction-tuned LLMs, fully fine-tuned with RLHF, and with limited downstream evaluation. We extend this analysis to pre-trained and instruction-tuned models using parameter-efficient SFT and DPO.

\paragraph{Setup} To probe the impact of the informativeness and quality of the preference dataset on alignment and downstream performance, we compare the performance of training our models using the two preference datasets, HH-RLHF and BeaverTails, on their harmless and helpful splits. BeaverTails is supposedly more informative and of better quality than HH-RLHF, particularly for the harmlessness and safe prompts. For detailed analysis, we show comparisons using supervised fine-tuning (SFT) and DPO over the pre-trained Mistral-7b and instruction-tuned Mistral-7b-Instruct models. We also probe the impact of the dataset quality as the training progresses, analyzing the effect on the stability of the training.

\paragraph{Results and Observations} We compare the performance of the models across helpfulness and harmlessness when aligned using a preference dataset of lower quality with a higher quality dataset using SFT and DPO in Figure~\ref{fig:hh_vs_pb}. The helpfulness benchmarks indicate that a higher quality and more informative preference dataset leads to overall more helpful behavior of the aligned model across both the alignment methods. When using the helpful preference subsets, there are bigger gains on general-purpose NLP tasks (such as MMLU) and instruction-following benchmarks (such as Alpaca Eval) when SFT is used. This suggests that SFT is more sensitive to the dataset quality when the downstream task and preference dataset are of a similar nature. When using the orthogonal harmless preference for alignment, relatively lower quality datasets such as HH-RLHF lead to overfitting when using alignment methods like DPO and experience significant performance degradation. However, since BeaverTails ensures that the safe and harmless responses are informative, there is no degradation in performance when used with DPO.

The harmlessness benchmarks reveal that using a more informative and safer dataset makes the model less harmful. However, DPO leads to a significant degeneration of the model when BeaverTails is used with Mistral-7b. As Mistral-7b is not instruction-tuned, we hypothesize that this degradation might be due to the inability of the base model to effectively represent the reward for preference optimization, specifically for samples targeted towards more objective preferences such as harmlessness and safety compared to broader preferences such as helpfulness. We also observe that harmless alignment using higher-quality datasets is more faithful than lower-quality datasets, especially when using SFT.

\begin{figure}[!htb]
	\centering
	\includegraphics[width=\linewidth]{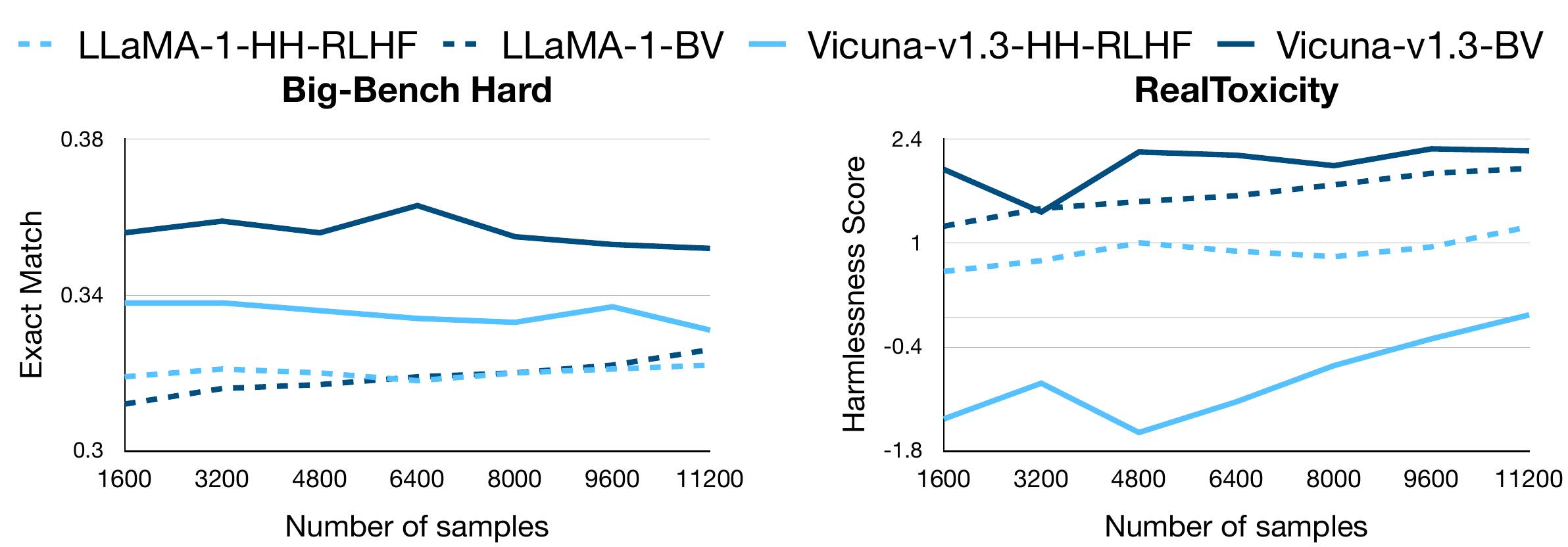}
	\caption{\label{fig:hh_vs_pb_samples} Performance trends w.r.t number of samples of HH-RLHf and BeaverTails used for SFT alignment (Section~\ref{subsec:hh_vs_bv}). Models aligned with a higher-quality dataset seem to learn faster or regress slower.}
\end{figure}

\begin{figure*}[!htb]
	\centering
	\includegraphics[width=\linewidth]{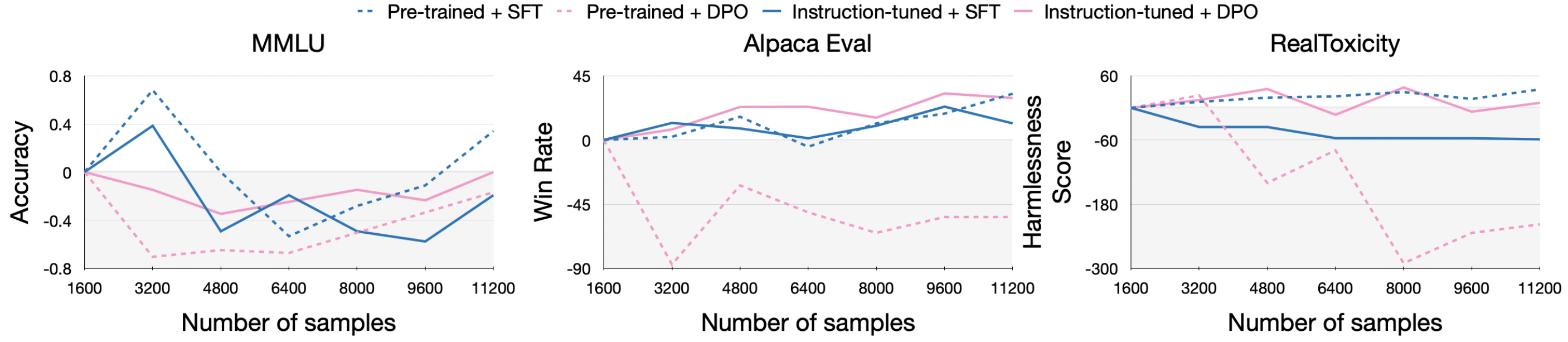}
	\caption{\label{fig:num_samples} Relationship of the number of samples used for alignment using SFT and DPO with Mistral (Section~\ref{subsec:num_samples}). The performance here is shown in \% relative to the performance when using 1600 samples.}
\end{figure*}

We also present the progress of the model performances at various training steps in Figure~\ref{fig:hh_vs_pb_samples}. We observe that when using a more informative preference dataset, the alignment either leads to faster learning and more gains or achieves comparable performance to the base model. When using a relatively lower-quality dataset, either the learning is slower or deterioration is observed during the course of the alignment training. Overall, better quality and informative datasets are better for alignment across methods and preferences when using PEFT.

\subsection{Impact of Quantity}
\label{subsec:num_samples}

To the best of our knowledge, no studies have investigated the relationship between the number of alignment samples used across methods, models, and alignment preferences.

\paragraph{Setup} We evaluate the relationship between the number of preference alignment samples used for SFT and DPO with the downstream performance. We use Mistral-7b and Mistral-7b-Instruct as the model due to their superior performance, and BeaverTails as our preferred dataset due to its more diverse samples. We evaluate the performance at multiple training steps over helpfulness and harmlessness benchmarks.

\paragraph{Results and Observations} Figure~\ref{fig:num_samples} indicates that instruction-tuned models are more robust to further alignment when using SFT. Furthermore, since the preference data does not directly contain examples of the downstream tasks, the performance of the models is reduced. It is interesting to note here that Mistral-7b-Instruct performs worse than the base model on MMLU, indicating that continued supervised fine-tuning is sensitive to the dataset samples used for training, particularly for general-purpose NLP tasks.

However, using DPO over the instruction-tuned models for instruction-following tasks gives consistent improvement, which aligns with previous works applying DPO as an alignment method~\citep{rafailov2023direct}. Models trained with DPO are also much more faithful to the preferences they are aligned with, particularly for harmlessness. This might be due to harmlessness being a relatively objective preference compared to helpfulness, which is much broader. However, as observed previously, the pre-trained models are sensitive to the preference alignment used over DPO, and since they are not strong inherent reward models, using DPO leads to a considerable degradation in their performance. A combination of SFT-based instruction tuning followed by DPO would make up for the optimal training strategy in case of alignment to explicit preferences such as safety and harmlessness. 

Overall, instruction-tuned models are robust to additional samples when aligned with SFT and often regress but benefit from DPO. Furthermore, instruction-tuned models might require fewer samples to adapt to preferences compared to pre-trained models, as more samples might degrade their performance on general-purpose tasks. Pre-trained models generally get better at alignment and instruction-following both. Accordingly, there might be a sweet spot in the number of samples used for alignment.

\begin{table*}[!htb]
	\centering
	\small
	\begin{tabular}{@{}llcccccc@{}}
\toprule
\multicolumn{1}{l}{\textbf{Alignment Method}} &                                      & \multicolumn{3}{c}{SFT}             & \multicolumn{3}{c}{DPO}             \\ \cmidrule(lr){3-5} \cmidrule(lr){6-8} 
\textbf{Model}                                & \multicolumn{1}{c}{\textbf{Variant}} & MMLU  & BBH   & RealToxicity & MMLU  & BBH   & RealToxicity \\ \midrule
\multirow{4}{*}{\textbf{Mistral-7b}}          & Original                             & \textbf{0.590} & 0.395 & 1.255               & 0.590 & 0.395 & \textbf{1.255}               \\
                                              & Harmless                             & 0.583 & 0.409 & \textbf{1.863}               & 0.590 & 0.402 & -3.701              \\
                                              & Helpful                              & \textbf{0.589} & \textbf{0.417} & 1.571               & \textbf{0.594} & \textbf{0.418} & -9.375              \\
                                              & Harmless+Helpful                     & 0.575 & 0.408 & 1.239               & 0.593 & 0.412 & -9.419              \\ \midrule
\multirow{4}{*}{\textbf{Mistral-7b-Ins}}      & Original                             & \textbf{0.535} & \textbf{0.385} & 1.273               & \textbf{0.535} & 0.385 & 1.273               \\
                                              & Harmless                             & 0.518 & 0.366 & \textbf{1.981}               & 0.529 & \textbf{0.392} & \textbf{3.520}               \\
                                              & Helpful                              & 0.519 & 0.371 & 1.145               & \textbf{0.535} & 0.382 & 2.170               \\
                                              & Harmless+Helpful                     & 0.518 & 0.367 & 0.991               & 0.531 & 0.374 & -0.025              \\ 

\bottomrule
\end{tabular}
	\caption{\label{tab:mixture_pref}Effect of aligning on a mixture of two distinct preferences (Harmlessness and Helpfulness) compared to training on individual preferences (Section~\ref{subsec:mixture_pref}). The model aligned to a mix of both preferences generally performs worse than the models aligned to the individual preferences.}	
\end{table*}

\begin{figure*}[!htb]
	\centering
	\includegraphics[width=\linewidth]{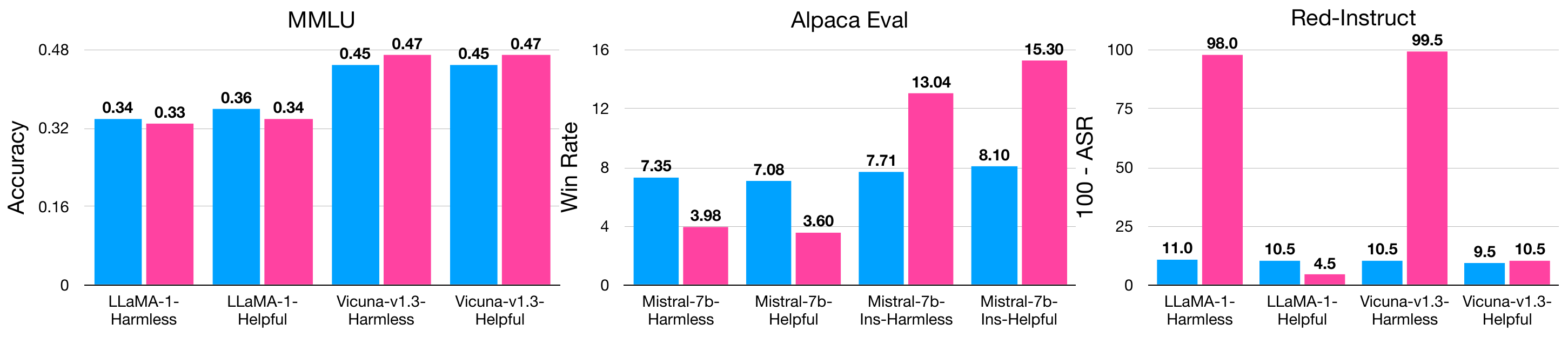}
	\caption{\label{fig:sft_vs_dpo} Comparing the downstream performance when aligning using SFT (in \textbf{\colorbox{brightcerulean!30}{light blue}}) and DPO (in \textbf{\colorbox{pink!60}{pink}}) with QLoRA. SFT outperforms DPO generally when used over pre-trained models, significantly for instruction following tasks. DPO is more faithful to explicit preferences such as harmlessness and performs significantly better for instruction-tuned models (Section~\ref{sec:sft_vs_dpo}).}
\end{figure*}

\subsection{Impact of Data Mixtures}
\label{subsec:mixture_pref}

Previous studies have shown that optimizing for individual preferences can result in trade-offs in performance~\citep{bai2022training}. We extend this study and evaluate models trained on a mixture of distinct preferences with PEFT against those trained on individual preferences.

\paragraph{Setup} We use both SFT and DPO to align our models on the harmless and helpful preferences of BeaverTails. We then combine both the preference sets and align models on a mix of the preferences for our comparison.

\paragraph{Results and Observations} From Table~\ref{tab:mixture_pref}, we see that when both SFT and DPO are used to align the models using a mixture of preferences, the model performs worse than models trained on individual preferences. This trend is consistent across both pre-trained and instruction-tuned models across the preferences. This might arise because the model encounters conflicting responses for prompts that might be similar to each other when orthogonal preferences are used for alignment, leading to non-optimal training. We observe larger degradations when using DPO with a mixture of the preferences compared to SFT, suggesting that DPO is more sensitive to the \textit{type} of samples used for alignment and requires more uniform preference samples. Overall, better curation of dataset mixtures when using alignment methods is necessary to achieve optimal performance.

\subsection{Key Takeaways}

\begin{itemize}
    \item Higher quality and more informative datasets lead to better alignment when using both SFT and DPO, with more significant gains when using SFT.
    \item Performing SFT on strong instruction-tuned models might not lead to gains as there is performance saturation, and it can even lead to degradation depending on the dataset.
    \item Training on a mixture of diverse preferences often leads to performance trade-offs and degradation across them.
\end{itemize}

\section{Effects of the Alignment Method}
\label{sec:sft_vs_dpo}

Previous works have shown that preference optimization methods, such as RLHF and DPO, are better than methods like SFT for full fine-tuning of models on standard preference datasets~\citep{christiano2023deep, rafailov2023direct}. We validate this claim using PEFT across the different preferences.

\begin{figure*}[!htb]
	\centering
	\includegraphics[width=\linewidth]{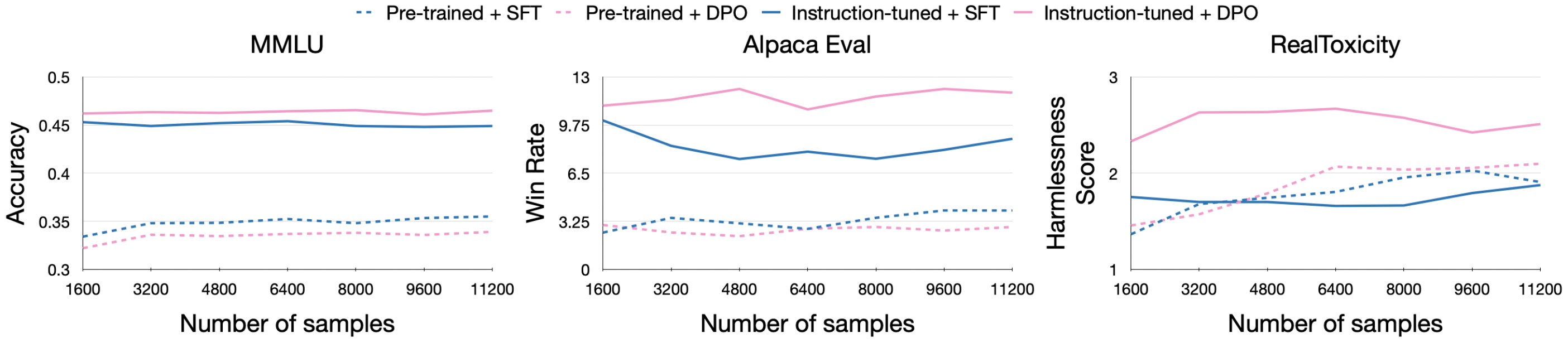}
	\caption{\label{fig:pretrain_vs_instruction} Comparing the effect of applying alignment methods on pre-trained models with instruction-tuned models using LLaMA-1 (Section~\ref{subsec:pretrain_vs_instruction}). SFT helps more for pre-trained models, while DPO helps more for instruction-tuned models. However, when aligning to objective preferences like harmlessness, DPO leads to more faithful alignment across both pre-trained and instruction-tuned models.}
\end{figure*}

\paragraph{Setup} We compare the performance attained when using SFT and DPO to train our models across preferences. We use all four models in our study to align with the harmless and helpful preferences of the BeaverTails dataset.

\paragraph{Results} We present the results comparing SFT and DPO in Figure~\ref{fig:sft_vs_dpo}. We observe that, on average, DPO leads to better performance when aligned with the helpful preferences compared to SFT for instruction-tuned models. This might stem from the fact that DPO takes into consideration both the chosen and the rejected samples and is able to extract more learnings from the preference dataset. When evaluating the harmlessness, we observe very interesting results. Using DPO is significantly better than SFT, indicating that DPO makes the model alignment much more faithful to the preferences than SFT. This is more pronounced when the preferences are more objective, such as harmlessness, compared to relatively broader preferences like helpfulness. This observation also holds for models such as LLaMA-1, where DPO performs worse than SFT on helpfulness benchmarks. Furthermore, it is not always the case that aligning using the harmless preference dataset leads to more harmlessness in the case of SFT, further showing that it is not as faithful as DPO. Given the inability of pre-trained models to inherently provide rewards for broader preferences, we can say that generally, if the base model can be expected to be a good reward model, using DPO leads to better downstream performance and more faithfully aligned models.

We also make another interesting observation that contradicts logical deductions. Oftentimes, using harmless data performs better than helpful data. This might be a characteristic of the BeaverTails dataset as it specifically contains helpful and safe responses, but this should be an important consideration when curating datasets for preference alignment.

\subsection{Key Takeaways}

\begin{itemize}
    \item DPO performs better than SFT and is more faithful to explicit preferences, such as harmlessness, compared to broader preferences, such as helpfulness.
    \item It is beneficial to instruction-tune models first with SFT and then apply DPO.
\end{itemize}

\section{Effects of the Nature of Base Models}
\label{sec:base_model}

\subsection{Contrasting Pre-trained and Instruction-Tuned Models}
\label{subsec:pretrain_vs_instruction}

Most of the works using DPO for alignment use it over instruction-tuned models. We perform a study comparing the effects of applying alignment methods over pre-trained and instruction-tuned models.

\paragraph{Setup} We investigate the effect of the nature of the underlying model trained for alignment to different preferences on the downstream tasks. We consider LLaMA-1 as the pre-trained model and Vicuna-v1.3 as the instruction-tuned variant. We compare the downstream performance when the models are aligned with SFT and DPO using the harmless and helpful preferences of BeaverTails.

\begin{table*}[!htb]
	\centering
	\small
	\begin{tabular}{lcccccc}
		\toprule
		\multirow{2}{*}{\textbf{Merge Variant}} & \multicolumn{3}{c}{\textbf{Mistral-7b}} & \multicolumn{3}{c}{\textbf{Mistral-7b-Ins}} \\ \cmidrule(lr){2-4} \cmidrule(lr){5-7} 
		             & \textbf{BBH}   & \textbf{Alpaca Eval} & \textbf{RT}    & \textbf{MMLU}  & \textbf{Alpaca Eval} & \textbf{RT}    \\ \midrule
		Harmless-SFT & 0.409          & 7.346                & \textbf{1.863} & 0.518          & 7.711                & 1.981          \\
		Helpful-SFT  & 0.417          & 7.081                & 1.571          & 0.519          & 8.095                & 1.145          \\
		Harmless-DPO & 0.402          & 3.975                & -3.701         & 0.529          & 13.043               & 3.520          \\
		Helpful-DPO  & 0.418          & 3.602                & -9.375         & 0.535          & 15.299               & 2.170          \\ \midrule
		Slerp-SFT    & 0.415          & 2.236                & -9.115         & 0.530          & 8.966                & 3.311          \\
		Slerp-DPO    & 0.425          & \textbf{7.721}       & 1.604          & 0.524          & 8.209                & 1.873          \\
		DARE-SFT     & 0.419          & 7.472                & 1.870          & 0.525          & 8.209                & 2.207          \\
		DARE-DPO     & 0.428          & 2.981                & -8.602         & 0.534          & \textbf{15.702}      & \textbf{3.717} \\
		DARE-SFT+DPO & \textbf{0.432} & 6.468                & 1.755          & \textbf{0.538} & 9.363                & 2.938          \\ \bottomrule
	\end{tabular}
	\caption{\label{tab:merge_models}Performance of merging models using different methods with the models trained on individual preferences (Section~\ref{subsec:merge}). Model merging variants perform better on average than individually aligned models.}
\end{table*}

\paragraph{Results and Observations} We present the results of aligning the pre-trained models and the instruction-tuned models with SFT and DPO in Fig~\ref{fig:pretrain_vs_instruction}. Considering SFT, we observe that when applied over instruction-tuned models, it generally leads to degradation of performance when using both the helpful and harmless preferences. This might be because the models being aligned are already robust due to extensive instruction-tuning using datasets similar to the datasets used for downstream evaluation. Since BeaverTails is different from downstream datasets, models aligned on the dataset regress in performance. However, when we align the pre-trained models, we observe decent improvements in the performance of the downstream tasks. This is in line with studies showing that instruction-tuning, especially on the helpful preferences, makes the model follow the instructions of the downstream task more effectively.

When aligning the pre-trained and instruction-tuned models with DPO, we observe that the alignment stability depends on the base model itself, as it also acts like a reward model. For strong base models, such as the instruction-tuned models, DPO leads to better gains. These gains are more significant for instruction-following tasks like Alpaca Eval. However, for weaker base models like LLaMA-1, using DPO leads to degradation in performance in contrast to SFT. Hence, instruction-tuned models overall seem to fare more coherently and are more faithful to the preferences when DPO is used, but with relatively weaker base models, an initial stage of instruction-tuning using SFT followed by DPO would be more beneficial.

\subsection{Effect of using Merged Models}
\label{subsec:merge}

Model merging methods via interpolations have recently proven to be effective~\citep{yadav2023tiesmerging}. We investigate the effects of merging models that are aligned on distinct preferences and trained with different methods.

\paragraph{Setup} To study the effect of various combinations of merging models across the preferences and alignment methods, we experiment with two recent merge methods: Spherical interpolation (Slerp)~\citep{10.1145/325165.325242} and DARE~\citep{yu2023language}. We experiment with merging models aligned on the harmless and helpful preferences using DPO and SFT (for example, Slerp-SFT refers to the merging of harmless and helpful models aligned with SFT). As DARE allows merging any number of models, we also experiment with merging all four aligned models, i.e. harmless and helpful alignments using SFT and DPO together (DARE-SFT+DPO).

\paragraph{Results and Observations} From Table~\ref{tab:merge_models}, we see that models aligned to preferences with SFT and DPO seem to perform better than the individual models when merged using DARE TIES, while merging with Slerp leads to performance degradation. The best performance is obtained when we merge all four models, which is very interesting. Overall, the performance of the merged model depends on the individual models, with the performance of strong models regressing when merged with very weak models. We leave this exploration of using elaborate merging methods as future work.

\subsection{Key Takeaways}

\begin{itemize}
    \item Pre-trained models generally align better with SFT, whereas instruction-tuned models align better with DPO.
    \item Instruction-tuned models have reduced performance if the SFT dataset is not similar to the evaluation task.
    \item Merged models can effectively mitigate performance trade-offs when aligning to diverging preferences.
\end{itemize}

\begin{figure*}[!htb]
	\centering
	\includegraphics[width=\linewidth]{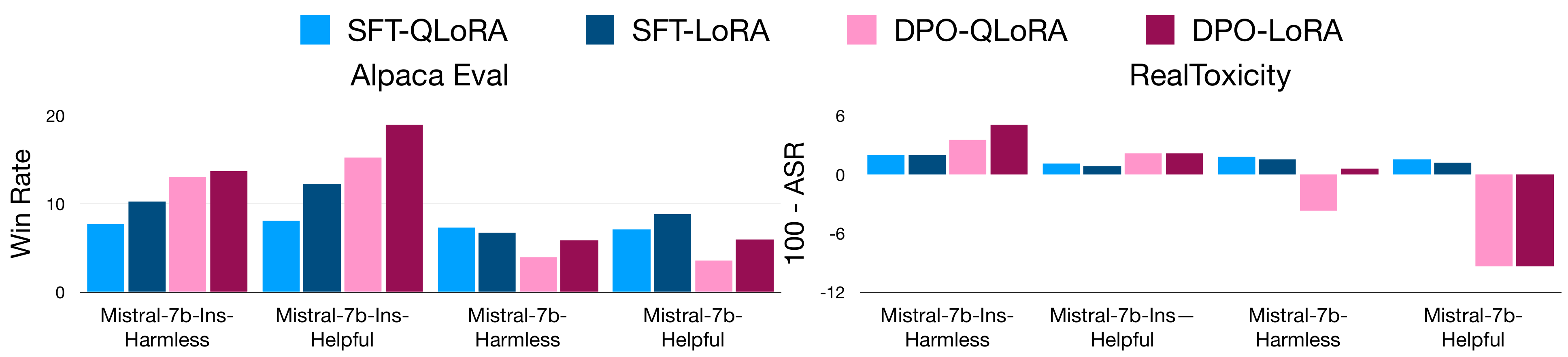}
	\caption{\label{fig:qlora_vs_lora} Comparing the performance when using QLoRA and LoRA as the PEFT method (Section~\ref{sec:qlora_vs_lora}). We observe similar trends and performance across the two methods.}
\end{figure*}

\section{Comparing QLoRA with LoRA}
\label{sec:qlora_vs_lora}

We compare the performance trends observed when using QLoRA as the PEFT method with LoRA as the PEFT method in Figure \ref{fig:qlora_vs_lora}. We observe similar patterns as observed when using QLoRA across the three axes of our study, and also observe similar performance as expected. We provide the extended results of using QLoRA and LoRA across the training setup in the Appendix \ref{sec:extended_results}.

\section{Key Takeaways}

Consolidating our findings and conclusions, we present key takeaways across the three core axes of preference alignment in PEFT settings.
\begin{itemize}
	[itemsep=2pt,parsep=2pt,partopsep=2pt,leftmargin=*,topsep=2pt]
	\item \textbf{Alignment Dataset}: Higher quality and informative datasets lead to better alignment when using both SFT and DPO, with more significant gains when using SFT. However, for instruction-tuned models, there might be performance saturation on certain tasks, and only a few samples might be required to gain improvements with SFT. Furthermore, in certain cases, aligning on high-quality distinct preferences can lead to improvements compared to aligning on the same preference.
	\item \textbf{Alignment Method}: DPO performs better than SFT and is more faithful to explicit preferences, such as harmlessness, compared to broader preferences, such as helpfulness. However, it fails or overfits when applied over pre-trained models. We suggest instruction-tuning models first with SFT and then apply DPO.
	\item \textbf{Nature of Base Model}: Pre-trained models generally align better with SFT, whereas instruction-tuned models align better with DPO. Instruction-tuned models have reduced performance if the SFT dataset is not similar to the evaluation task. Models merged across distinct preferences generally perform better than the models aligned on individual preferences.
\end{itemize}

\section{Conclusion and Future Work}

In this work, we perform an extensive analysis of the trade-offs of downstream performance of models over distinct preferences across three core alignment axes: the alignment dataset, the preference alignment method, and the nature of the base model, particularly when using PEFT. We use the two most widely-used 7 billion parameter models, LLaMA-1 and Mistral-7b, along with their instruction-tuned variants, two popular preference alignment datasets, HH-RLHF and BeaverTails, two commonly used alignment methods, SFT and DPO, and evaluate on $5$ benchmarks across helpfulness and harmlessness. By conducting over 300 experiments with QLoRA and LoRA, our findings reveal interesting trends that address the limitations of existing literature and often deviate from them. We consolidate our findings into key takeaways and hope that our guidelines will help researchers perform more effective parameter-efficient preference alignment.

As future work, we aim to extend our study on trade-offs across multiple preferences, spanning various domains for alignment. We also plan to explore PEFT methods, such as model merging and mixture-of-experts, to mitigate these trade-offs.

\section*{Limitations}

We acknowledge the limitations of our work:
\begin{itemize}[itemsep=2pt,parsep=2pt,partopsep=2pt,leftmargin=*,topsep=2pt]
	\item We selected LoRA and QLoRA as parameter-efficient training methods for our experiments because of their relatively small performance degradation over massive compute savings. We repeatedly point this out throughout our work to avoid confusion; however, other PEFT methods exist, such as adapters and prompt- and prefix-tuning.
	\item Since we focus on analyzing settings accessible to diverse researchers, we focus on models with 7B parameters. Our findings may or may not extend to models of smaller or larger sizes.
	\item We conduct experiments to identify the trade-offs across only two preferences. Based on existing literature, we use the most widely used preferences, harmlessness and helpfulness. However, the findings might not extend when more than two preferences or domains are involved for alignment. These more complicated settings might require a separate in-depth study.
\end{itemize}

\section*{Ethics Statement}

Our research studies various LLMs, parameter-efficient methods, and datasets for preference alignment. We recognize that ensuring the safety of LLMs is a crucial concern and believe that they must be thoroughly vetted before being deployed. Our investigation aims to assist the diverse community of researchers and individuals seeking to align models for safety. Such a research direction is especially important since each person and community has their own perception of safety.

We also acknowledge that training and evaluating LLMs for safety is a sensitive issue. No method can guarantee the complete safety of models, and a comprehensive evaluation of models should be carried out even after using the techniques and results described in our work before deploying them in real-world human-facing situations.

\section*{Acknowledgements}

Sarath Chandar is supported by the Canada CIFAR AI Chairs program, the Canada Research Chair in Lifelong Machine Learning, and the NSERC Discovery Grant. The project was also supported by the IBM-Mila collaboration grant. The authors acknowledge the computational resources provided by the Digital Research Alliance of Canada. The authors would also like to Shravan Nayak for his inputs and experimental analysis.

\bibliography{main}
\bibliographystyle{acl_natbib}

\clearpage
\appendix

\input{appendix}

\end{document}

%% file: appendix.tex
\section{Extended Results}
\label{sec:extended_results}

\begin{table*}[!htb]
	\centering
	\small
	\setlength{\tabcolsep}{6pt}
	\begin{tabular}{@{}llcccccccc@{}}
		\toprule
		& Evaluation Dataset                  & \multicolumn{2}{c}{BBH}               & \multicolumn{2}{c}{MMLU} & \multicolumn{2}{c}{AlpacaEval}         & \multicolumn{2}{c}{RT} \\ \cmidrule(lr){3-4}\cmidrule(lr){5-6}\cmidrule(lr){7-8}\cmidrule(lr){9-10} 
		\multirow{-2}{*}{Alignment Dataset} & Model                   & SFT   & DPO   & SFT   & DPO   & SFT   & DPO    & SFT   & DPO    \\ \midrule
		                                    & Mistral-7b-Ins-Harmless & 0.356 & 0.378 & 0.507 & 0.527 & 2.239 & 4.738  & 0.533 & 2.662  \\
		                                    & Mistral-7b-Ins-Helpful  & 0.354 & 0.380 & 0.502 & 0.530 & 2.488 & 8.853  & 0.563 & 1.907  \\
		                                    & Mistral-7b-Harmless     & 0.402 & 0.149 & 0.554 & 0.484 & 0.871 & 0.498  & 0.882 & 3.199  \\
		                                    & Mistral-7b-Helpful      & 0.408 & 0.397 & 0.571 & 0.583 & 2.369 & 3.095  & 1.055 & 3.234  \\
		                                    & LLaMA-1-Harmless        & 0.316 & 0.223 & 0.302 & 0.316 & 0.249 & 0.248  & 1.222 & 2.381  \\
		                                    & LLaMA-1-Helpful         & 0.322 & 0.280 & 0.338 & 0.325 & 0.373 & 2.224  & 1.224 & 1.352  \\
		                                    & Vicuna-v1.3-Harmless    & 0.314 & 0.331 & 0.419 & 0.460 & 2.236 & 9.559  & 0.033 & 4.682  \\
		\multirow{-8}{*}{HH-RLHF}           & Vicuna-v1.3-Helpful     & 0.331 & 0.335 & 0.435 & 0.456 & 3.731 & 8.577  & 1.639 & 1.884  \\ \midrule
		                                    & Mistral-7b-Ins-Harmless & 0.366 & 0.392 & 0.518 & 0.529 & 7.711 & 13.043 & 1.981 & 3.520  \\
		                                    & Mistral-7b-Ins-Helpful  & 0.371 & 0.382 & 0.519 & 0.535 & 8.095 & 15.299 & 1.145 & 2.170  \\
		                                    & Mistral-7b-Harmless     & 0.409 & 0.402 & 0.583 & 0.590 & 7.346 & 3.975  & 1.863 & -3.701 \\
		                                    & Mistral-7b-Helpful      & 0.417 & 0.418 & 0.589 & 0.594 & 7.081 & 3.602  & 1.571 & -9.375 \\
		                                    & LLaMA-1-Harmless        & 0.321 & 0.304 & 0.341 & 0.328 & 2.857 & 1.370  & 2.008 & 3.364  \\
		                                    & LLaMA-1-Helpful         & 0.326 & 0.311 & 0.355 & 0.339 & 3.975 & 2.864  & 1.907 & 2.098  \\
		                                    & Vicuna-v1.3-Harmless    & 0.345 & 0.358 & 0.453 & 0.468 & 9.689 & 9.955  & 2.246 & 4.711  \\
		\multirow{-8}{*}{BeaverTails}       & Vicuna-v1.3-Helpful     & 0.348 & 0.355 & 0.449 & 0.465 & 8.820 & 11.940 & 1.876 & 2.510  \\ \bottomrule
	\end{tabular}
	\caption{\label{tab:extended_results} Results across our evaluation datasets when trained with different alignment methods and alignment datasets using QLoRA.}
\end{table*}

\begin{table*}[!htb]
	\centering
	\small
	\begin{tabular}{@{}lcccccccc@{}}
\toprule
Evaluation Dataset                  & \multicolumn{2}{c}{BBH}               & \multicolumn{2}{c}{MMLU} & \multicolumn{2}{c}{AlpacaEval}         & \multicolumn{2}{c}{RT} \\ \cmidrule(lr){2-3}\cmidrule(lr){4-5}\cmidrule(lr){6-7}\cmidrule(lr){8-9} 
                        & SFT        & DPO        & SFT         & DPO        & SFT                            & DPO    & SFT       & DPO        \\ \midrule
Mistral-7b-Ins-Harmless & 0.363      & 0.384      & 0.515       & 0.522      & 10.323                         & 13.699 & 1.975     & 5.117      \\
Mistral-7b-Ins-Helpful  & 0.373      & 0.379      & 0.517       & 0.529      & 12.313                         & 19.006 & 0.862     & 2.206      \\
Mistral-7b-Harmless     & 0.414      & 0.413      & 0.582       & 0.585      & 6.733                          & 5.853  & 1.587     & 0.641      \\
Mistral-7b-Helpful      & 0.419      & 0.415      & 0.584       & 0.594      & 8.864                          & 5.970  & 1.261     & -9.369     \\
LLaMA-1-Harmless        & 0.312      & 0.308      & 0.342       & 0.323      & 4.608                          & 0.994  & 1.739     & 3.204      \\
LLaMA-1-Helpful         & 0.307      & 0.320      & 0.354       & 0.330      & 4.857                          & 2.989  & 1.655     & 1.588      \\
Vicuna-v1.3-Harmless    & 0.342      & 0.332      & 0.450       & 0.454      & 11.180                         & 13.292 & 2.007     & 5.367      \\
Vicuna-v1.3-Helpful     & 0.343      & 0.335      & 0.451       & 0.453      & 12.422 & 20.449 & 1.703     & 3.231      \\ \bottomrule
\end{tabular}
	\caption{\label{tab:extended_results_lora} Results across our evaluation datasets when trained with different alignment methods using BeaverTails with LoRA.}
\end{table*}

We present our results across the benchmarks when the four models are trained on HH-RLHF and BeaverTails using SFT and DPO with QLoRA in Table \ref{tab:extended_results} and with LoRA in Table \ref{tab:extended_results_lora}. We observe that the trends presented in the main work hold across preference alignment datasets, alignment methods, and base models.

\section{Hyperparameter settings}

\begin{table*}[!htb]
	\centering
	\small
	\setlength{\tabcolsep}{5pt}
	\begin{tabular}{@{}lcc@{}}
		\toprule
		                        & SFT             & DPO                                                                  \\ \midrule
		Learning Rate           & 2.00E-04        & 5.00E-05                                                             \\
		PEFT Rank               & 64              & 16                                                                   \\
		PEFT Alpha              & 16              & 32                                                                   \\
		PEFT Dropout            & 0.1             & 0.05                                                                 \\
		Batch size              & 16              & 16                                                                   \\
		Max Grad Norm           & 0.3             & 1                                                                    \\
		Weight Decay            & 0.001           & 0                                                                    \\
		Optimizer               & Paged AdamW     & Paged AdamW                                                          \\
		Learning Rate Scheduler & Constant        & cosine                                                               \\
		Adam Beta1              & 0.9             & 0.9                                                                  \\
		Adam Beta2              & 0.999           & 0.999                                                                \\
		PEFT Target Modules     & q\_proj,v\_proj & k\_proj, gate\_proj, v\_proj, up\_proj, q\_proj, o\_proj, down\_proj \\ \bottomrule
	\end{tabular}
	\caption{\label{tab:hparams}Hyperparameter settings for our alignment methods.}
\end{table*}

For LoRA and QLoRA parameters, we do a grid search with different combinations of LoRA and QLoRA rank and alpha. For SFT, we experiment with ranks $[32,64,128]$ and alphas $[16,32,64]$. For DPO, we experiment with ranks $[8,16,32]$ and alphas $[16,32,64]$.
We choose the best hyperparameters based on the validation set performance of the alignment datasets.
We show the detailed best hyperparameter setup of our training setup during alignment training in Table \ref{tab:hparams}.

\section{Alignment Dataset Creation and Details}

The HH-RLHF dataset has explicit splits for harmlessness and helpfulness. However, BeaverTails does not have this explicit split. Since the BeaverTails dataset has no explicit ``helpful'' and ``harmless'' splits like HH-RLHF, we create these splits using the provided labels. Each sample in the dataset has an instruction and two responses. Each response also has a ``'safe'' label, and the dataset sample has the label of the ``better'' response. To make helpfulness alignment data, we select the "better" response as the choice for helpfulness and select the second response as the "rejected" sample. Similarly, to create harmlessness alignment data, we select the samples where the response was marked safe and the second response was marked unsafe.

The harmless split of HH-RLHF has $42507$ samples, and the helpful split has $43810$ samples for alignment training. The samples are multi-turn. Using our method, the helpful split of BeaverTails has $297394$ samples, and the harmless split has $46625$ samples, and the samples are single-turn. We modify the prompts and outputs for both datasets following the suggested instruction-tuning templates of the Vicuna and Mistral models.

\noindent \textbf{Vicuna:} \texttt{A chat between a curious user and an artificial intelligence assistant. The assistant gives helpful, detailed, and polite answers to the user's questions. \\
	USER: ``prompt'' \\
ASSISTANT: ``response''</s>}

\noindent \textbf{Mistral:} \texttt{[INST] ``prompt'' [/INST] ``response'' </s>}

\section{Evaluation Dataset Statistics}

\paragraph{MMLU} The MMLU dataset has $56168$ samples across 57 tasks.

\paragraph{Big-Bench Hard} The BBH dataset has $6511$ samples across 23 tasks.

\paragraph{Alpaca Eval} The Alpaca Eval dataset has $805$ human written prompts.

\paragraph{RealToxicity} The full RealToxicity prompts dataset has 99.4k prompts. We use the $1000$ most severe prompts for our evaluation.

\paragraph{Red-Instruct} We use the DangerousQA set of the Red-Instruct dataset, which has 200 prompts. We also use the Chain-of-Utterance format, which is the strongest in the dataset.

\section{Model and Compute Statistics}

All our model variants are the 7-billion parameter versions of the original models. For our experiments, we use a $40$GB A100 GPU. SFT alignment takes about $15$ minutes for BeaverTails and $20$ minutes for HH-RLHF. DPO alignment takes about $60$ minutes for BeaverTails and $90$ minutes for HH-RLHF.

\section{Background and Related Work}

\subsection{Alignment Methods}

Alignment training aims to reduce the mismatch between an LLM's pre-training and user preference requirements. It also ensures that models are safe and harmless, reducing the risks associated with their use. We choose the two most widely used alignment methods:

\paragraph{Supervised fine-tuning (SFT)} SFT uses a pair of input instructions and corresponding gold answers or outputs to fine-tune the LLM using autoregressive language modeling. The training objective is similar to pre-training, but the dataset is orders of magnitude smaller and follows a strict format. This method is often used for the 'instruction-tuning' stage for models like Alpaca~\citep{alpaca} and Mistral-7b-Instruct~\citep{jiang2023mistral}.

\paragraph{Reinforcement Learning from Human Feedback (RLHF)} RLHF~\citep{ziegler2019finetuning} is a reinforcement learning-based alignment method that consists of three steps: 1. Collect human feedback data, 2. Train a reward model on the feedback data. 3. Fine-tune an LLM with RL using PPO~\citep{schulman2017proximal} and the reward model. RLHF is the most commonly used method for preference alignment but often requires a lot of computation and steps for alignment. Various variants of RLHF have been proposed, such as using pure RL for training LLMs with human feedback in an online manner~\citep{bai2022training} and modifying the reward modeling with adversarial probing~\citep{glaese2022improving}.

\paragraph{Direct Preference Optimization (DPO)} Reinforcement learning algorithms such as RLHF~\citep{christiano2023deep} have been used to better align LLMs using paired human preference data: data consisting of accepted and rejected outputs given an instruction. However, these methods require multiple steps as well as training of separate reward models. DPO~\citep{rafailov2023direct} is a method that inherently uses the model being aligned as a reward model, making alignment more stable and optimized.

\subsection{Parameter-Efficient Training (PEFT)}

Parameter-efficient training methods require updating only a fraction of the parameters of the full model while achieving performance close to that of full fine-tuning. 
Examples of parameter-efficient training methods include using adapters~\citep{houlsby2019parameter}, prefix-tuning~\citep{li-liang-2021-prefix}, and prompt-tuning~\citep{lester-etal-2021-power}. Low-Rank Adaptation (LoRA)~\citep{hu2022lora} is another popular method that performs on par with full fine-tuning. LoRA works by inserting a smaller number of new weights into the model, which are the only trainable parameters. These weights are essentially low-rank matrix decompositions of the different model parameters. 
Low-rank decomposition refers to the process of approximating a larger matrix as a product of smaller matrices by assuming that the two smaller matrices are representative of the larger matrix. Assuming a parameter of size $A \times B$ and a low-rank $n$, i.e., $n << A, B$. We can hope to approximate $A \times B$ using $A \times n$ and $n \times B$. In the former cases, the parameters to be updated are $A \times B$, whereas in the latter case, the parameters are $A \times n + n \times B$. If the approximation is sufficient, we only need to update this new set of parameters, and they can later be 'merged' with the main model after training. Quantized LoRA (QLoRA)~\citep{dettmers2023qlora} takes the efficiency of LoRA a step ahead by QLoRA by also quantizing the weights of the LoRA adapters (smaller matrices) to lower precision (e.g., 4-bit instead of 8-bit), reducing the memory footprint and storage requirements while getting similar performance.